\documentclass{acmsiggraph}               

\usepackage{graphicx}
\usepackage{url}
\usepackage{times}
\usepackage{amsmath}
\usepackage{algorithm}
\usepackage{algorithmic}

\title{Quantization of Deep Neural Networks to facilitate self-correction of weights on Phase Change Memory-based analog hardware}

\author{Arseni Ivanov\thanks{e-mail: ar3307iv-s@student.lth.se}
}
\affiliation{Independent Researcher\\ Sweden}


\begin{document}

\maketitle

\begin{abstract}
In recent years, hardware-accelerated neural networks have gained significant attention for edge computing applications. Among various hardware options, crossbar arrays, offer a promising avenue for efficient storage and manipulation of neural network weights. However, the transition from trained floating-point models to hardware-constrained analog architectures remains a challenge. In this work, we combine a quantization technique specifically designed for such architectures with a novel self-correcting mechanism. By utilizing dual crossbar connections to represent both the positive and negative parts of a single weight, we develop an algorithm to approximate a set of multiplicative weights. These weights, along with their differences, aim to represent the original network's weights with minimal loss in performance. We implement the models using IBM's aihwkit and evaluate their efficacy over time. Our results demonstrate that, when paired with an on-chip pulse generator, our self-correcting neural network performs comparably to those trained with analog-aware algorithms.
\end{abstract}

\section{Introduction}
An emerging area in neural network hardware is the analog compute paradigm. In order to get around the von-Neumann bottleneck, compute and memory is moved into a shared area, often implemented using crossbar arrays \cite{Lee2011}. This allows us to reduce the computational complexity of certain operations, such as Matrix-vector multiplication(MVM) from O($N^2$) to O(1) by utilizing properties of analog electronics with Kirchoffs laws.

\subsection{Background and Challenges}
In all current proposed variations of analog hardware, we find a certain weakness that causes there to be a trade-off between the device and required qualities for neural network implementation. Phase Change Memory(PCM) is a device variation which has shown promise in the field\cite{pcm_history}. As for the weaknesses with PCM devices, it is that they are susceptible to various kinds of noise. These are: write/programming noise, read noise, and weight/conductance drift. Write and read noise is applied when the respective action is performed on the analog weight, whilst the weight drift is tied to the inherent material properties of the PCM device.
A concise description of a PCM device can be found in aihwkit's documentaiton\cite{aihwkit}.
\begin{quote}
``A PCM device consists of a small active volume of phase-change material sandwiched between two electrodes. In PCM, data is stored by using the electrical resistance contrast between a high-conductive crystalline phase and a low-conductive amorphous phase of the phase-change material. The phase-change material can be switched from low to high conductive state, and vice-versa, through applying electrical current pulses. The stored data can be retrieved by measuring the electrical resistance of the PCM device.''
\end{quote}

These noise types can drive weights away from their intended values, leading to network inaccuracies. Existing techniques to counteract this include noise-aware training, differential weight representation, and global weight drift compensation.

\subsection{Our Contribution}

We propose a solution that combines a built-upon existing technique for differential weight representation, weight quantization as well as a novel self-correcting mechanism. Our algorithm minimizes the loss between the original and quantized weights by finding optimal quantization bins through simulated annealing. The self-correcting mechanism further ensures long-term network stability.


\section{Method}

\subsection{Theoretical setup}
\begin{figure}[h]
    \centering
    \includegraphics[scale=0.4]{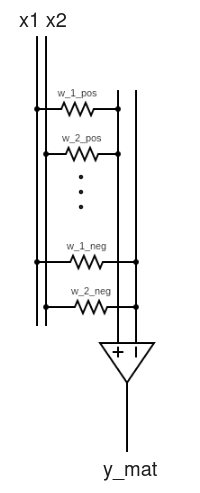}
    \caption{Simplified view of the two-element representation of two weights($w_1$ and $w_2$) and two inputs $x_1$ and $x_2$ creating a matrix multiplication output $y_{mat}$ by using the difference between the positive and negative lines.}
    \label{schema}
\end{figure}

We employ a two-element differential representation of each weight which we can visualize it in the simplified diagram in Figure \ref{schema}. In reality, we will also need source lines and converters between analog and digital. 

This structure has previously been employed in analog neural networks \cite{hardwareconv} as it reduces the effects of weight drift/perturbations that affects the hardware. If all weights are shifted 5 mV, a weight represented by a difference will stay the same. 

The inputs get sent to both the positive and negative weight for that input, which themselves accumulate onto the output line using Kirchoff's laws. All weights in the system are represented with positive resistances, which means that we can subtract the accumulated output from the negative output line from the positive one. This lets us have negative weights represented by positive numbers in the system, which often are required by neural networks to work efficiently.

\subsection{Network selection and training procedure}
Firstly, we need to select a problem and train a neural network to perform a task. For this experiment, we use the MNIST dataset, and a simple convolutional neural network that is chosen from a known architecture that was previously successfully implemented on crossbar arrays\cite{hardwareconv}.

We then impose some constraints during the training of the neural network. This includes adding weights below a value $\epsilon$ to the loss function. This discourages weights $w < \epsilon$, which would otherwise either require very small bins, or a very small difference between two bins in our architecture. Both of these are unwanted as the noise will affect those weights in a much larger proportion to their size. We can visualize the effect of this constraint in Figure \ref{weight_bins}. 

We also add a constraint to large weights above a value $\theta$. This is due to the conductance drift in the weights, which is larger the larger the weight is when using PCM-based crossbar arrays.
\subsection{Simulated Annealing for Bin Optimization}

Then, we perform simulated annealing to find the best bins for the task. The constraints for the optimization are as following:

\begin{itemize}
\item \textbf{Quantization Levels Constraint:} We should find two sets, one positive and one negative set. Each set should have \(N\) distinct quantization levels and together create a set of bins.
\item \textbf{Bin Constraint:} The possible bins in any found quantization set are given by \( SQ = \{d_{\text{pos}}, d_{\text{neg}}, (d_{\text{pos}_i} - d_{\text{neg}_j} \,|\, i \in d_{\text{pos}}, j \in d_{\text{neg}})\} \).
\item \textbf{Divisibility Constraint:} Each quantization level in a set must be divisible by the smallest factor in the set. They do not, however, have to be linearly distributed.
\item \textbf{SNR Constraint:} The step-multiple values $d[0]_{pos}$ and $d[0]_{neg}$ in the set should be larger than the write noise $\delta$ constraint which depends on the hardware and the programming procedure.
\item \textbf{Bin Difference Constraint:} The difference between the smallest positive and negative bins in the set ($abs(d[0]_{pos} - d[0]_{neg})$) must be larger than read noise error threshold $\epsilon$.
\end{itemize}

\begin{description}
    \item[\(N\):] The number of distinct quantization levels in each set.
    \item[\(d_{\text{pos}}\):] The set of positive quantization levels.
    \item[\(d_{\text{neg}}\):] The set of negative quantization levels.
    \item[\(\delta\):] The write noise constraint, which depends on the hardware and the programming procedure.
    \item[\(\epsilon\):] The read noise error constraint, which depends on the trade-off between write noise and read noise.
\end{description}

We provide details on the cost function, cooling schedule, and selection mechanism, showcasing how this approach leads to optimal bin selection.

The goal of the algorithm is to minimize the error between original weights, and the weights quantized using a found quantization set combination. Below is a pseudocode implementation of the algorithm:

\begin{algorithm}
\caption{Optimization of Bins Using Simulated Annealing}
\label{alg:sim_annealing}
\begin{algorithmic}[1]
\STATE \textbf{Input:} Neural net weights \( W \), positive and negative parts \( W_{\text{pos}} \), \( W_{\text{neg}} \), number of bins \( N \)
\STATE Initialize \( d_i \) for \( W \in \{W_{\text{pos}}, W_{\text{neg}}\} \)
\STATE Create quantization sets and calculate set \( SQ \)
\STATE Initialize current error, best error, and temperature \( T \)
\FOR{iteration \( i \) in range(iterations)}
\STATE Update temperature \( T \)
\STATE Perturb positive and negative bases
\STATE Propose new positive and negative bins
\STATE Compute error for proposed bins
\IF{proposed error $<$ current error \textbf{or} random value $<$ \(\exp\left(-\frac{{\text{proposed error} - \text{current error}}}{T}\right)\)}
\STATE Update current positive base, negative base, and error
\IF{proposed error $<$ best error}
\STATE Update best positive base, negative base, and error
\ENDIF
\ENDIF
\ENDFOR
\STATE Return best positive bins, negative bins
\end{algorithmic}
\end{algorithm}

It is possible to choose in step 7 in \textbf{Algorithm 1} if we want to enforce a linear constraint on the found bins such that any bin is a previous bin with the smallest factor $N[0]$ added. A linear constraint can simplify the search, but might not find the best result.

\subsection{Self-Correction Mechanism}
In our framework, we introduce a self-repairing mechanism that leverages the quantized weight levels to correct drifts in analog weight representations over time. The mechanism consists of four main components: an error threshold, a correction condition, a weight identification process, and an on-chip correction methodology.

\subsubsection{Error Threshold}
To quantify the deviation in the network's state, we define an error threshold based on the modulus of the weight values. Specifically, if any weight value modulus grows beyond \( \frac{N}{3} \) of its initial quantized level, the weight is considered a candidate for adjustment. Here, \( N \) is the quantization level multiple that was used initially for that specific weight. The error threshold comes with an power/accuracy trade-off. If we wait too long with re-adjusting bins, the weight might drift to an extent where the closest multiple no longer is the initial multiple. This leads to an irreversible degradation in the overall network performance for the remainder of its operational lifetime as we will no longer be able to get the initial network values until we reset the weights using a different mechanism.

\subsubsection{Correction Condition}
The network-wide condition for triggering the self-correction mechanism is based on global error estimation. By periodically pulsing an identity matrix through the network and accumulating the outputs, we can compare the current state of each layer against a baseline recorded at \( t=0 \). If the sum of the absolute differences across all weights exceeds a pre-defined global threshold, the self-correction mechanism is triggered.

\subsubsection{Weight Identification}

Once the correction condition is met, we proceed to identify the weights contributing most to the drift. This is done by selecting groups of weights, for example a layer of weights, and comparing the identity matrix output with it's initial output at $t=0$. If we have exceeded a layer-based drift difference threshold $dt$, we move on to the correction. 

In some cases, it can be cheaper to just reprogram the entire network, but in other cases where we have noise-sensitive layers such as CNN's, it might be sufficient to only reprogram those.

\subsubsection{On-Chip Correction Methodology}
To correct the identified weights, we use short programming pulses to nudge them back to their original multiple-based states. The magnitudes and durations of these pulses are determined based on the difference between the current and target states of each weight, as well as the current magnitude of the weight. This can be performed by an on-chip pulse generator\cite{pulse_generator}.

\subsubsection{Advantages and Applications}
The self-correction mechanism enhances the network's resilience to hardware-induced drifts, thus making it more robust for long-term deployments in edge computing scenarios. Moreover, the mechanism opens the door to more aggressive quantization strategies, as minor errors introduced by quantization can be periodically corrected, further reducing the computational and storage overhead.

\subsection{Compression}
Another benefit of the chosen multiple-quantization is that we can efficiently apply compression techniques such as those used in weight clustering to the weights. We can represent the positive and negative layers with integer matrices in range [0,M] where M is the largest multiple-factor used. This allows us to use N-bit representations of the weights, more generally $2^N-1 < M$ of the value, such as 4-bit weight representations if M $<$ 16. 

The lower representation range of values yields more repetition in the weight matrices, and allows for more aggressive compression of the weights.

\subsection{Testing methodology}
\begin{figure}[h]
    \centering
    \includegraphics[width=1\columnwidth]{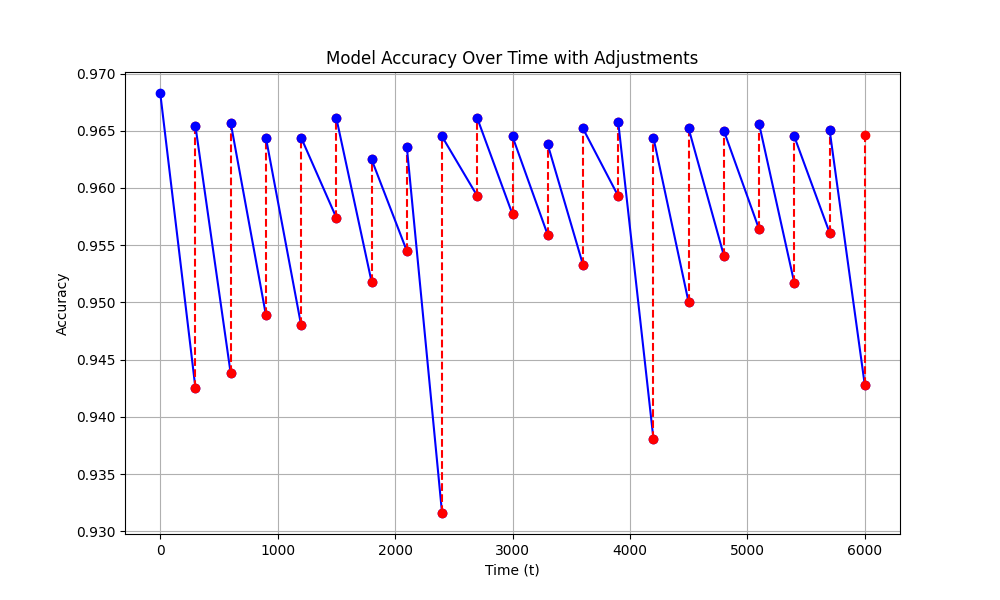}
    \caption{Digital weights over time with drift applied every 300 seconds. The red points signify accuracy after drift, while the blue points after a dotted red line signify the accuracy after adjustment.}
    \label{digital_weights}
\end{figure}

The accuracy of the self-repairing and the hardware-awarely trained networks are tested in time steps of 5 minutes. During every step, noise is added to the weights. At every timestep, the self-repairing neural network is probed for repair if a threshold of the cumulative layer error is exceeded. We compare the networks over 20 timesteps and note the accuracies in Figure \ref{digital_weights} and Digure \ref{analog_weights}.

\section{Results}
\begin{figure}[h]
    \centering
    \includegraphics[width=1\columnwidth]{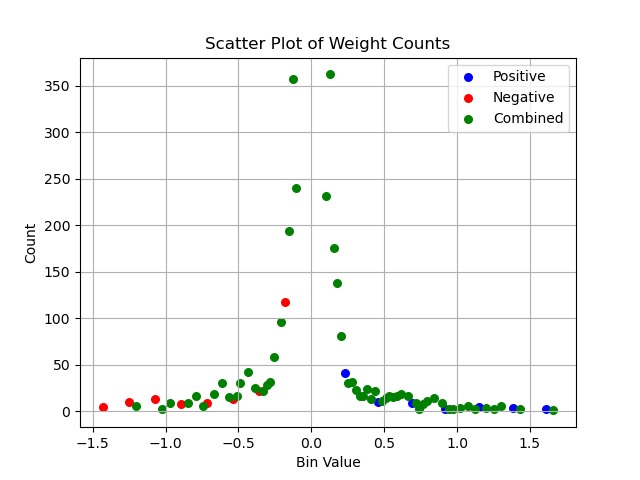}
    \caption{Scatter plot of amount of weights in each quantized bin in the set SQ with the best found quantization. Red dots signify negative weights, blue positive weights and green the weights defined using combinations of a positive and a negative weight.}
    \label{weight_bins}
\end{figure}

We train the candidate CNN network in a traditional fashion and achieve a f1-accuracy of 97.7\% on the MNIST dataset. We then apply the quantization and visualize the distribution of the weights in Figure \ref{weight_bins}. We can see that due to our constraints on the network weights enforced by the loss function, we find the first bins at $\epsilon$ distance away from 0. This quantization of weights keeps our initial accuracy of 97.7\%.

We then evaluate the network with 20 time-steps of 300 seconds drift each. At every time-step, we let the self-repairing network adjust it's weights into the closest positive and negative multiples. 

\begin{figure}[h]
    \centering
    \includegraphics[width=1\columnwidth]{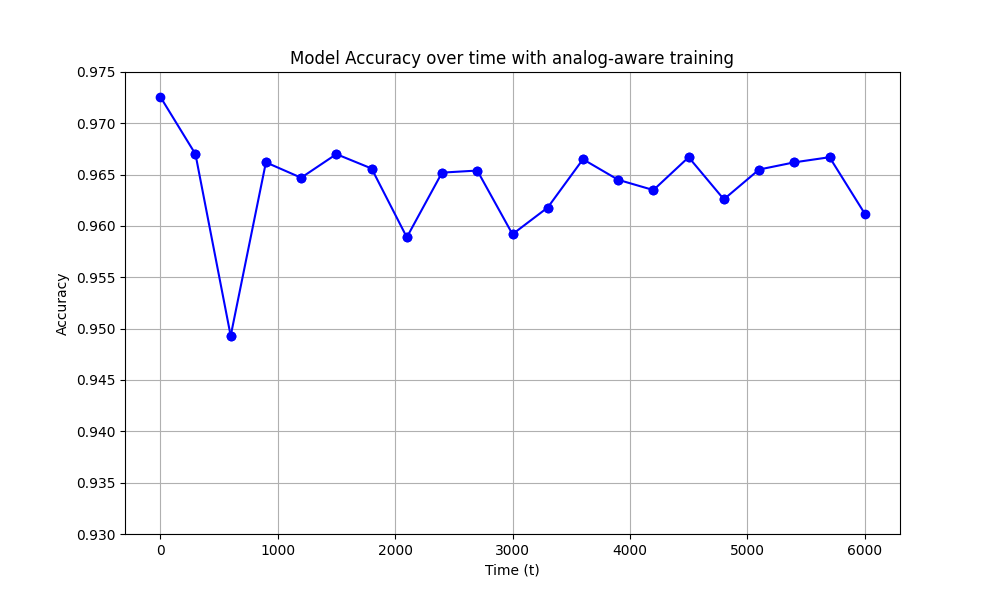}
    \caption{Analog-awarely trained weights over time with drift applied every 300 seconds.}
    \label{analog_weights}
\end{figure}

Alongside the self-repairing network, we train a hardware-aware analog neural network using the same network architecture and plot it's performance over the same timespan in Figure \ref{analog_weights}.

Both models were trained with the same analog noise configuration. The PCM noise configuration is given in Appendix \ref{code}.

\section{Discussion}
We find that the self-repairing network manages to keep it's accuracy stable once correcting itself, but in between the corrections it has a much wider variance of accuracy compared to the analog-awarely trained network. 

The constraint of weight being larger than $\delta$ allows us to represent small weights as a combination of a positive and negative weight. This is useful as shown by \cite{pulse_generator} where the proposed on-chip pulse generator has a significantly larger pulse error for smaller pulses. Pulses of size 100nA have up to 6\% average programming error, whilst pulses of 1.28mA have a 0.2\% average error.

Note that we do have to keep in mind that since we are working with small numbers, a high enough read noise error will mean that we will due to propagation of uncertainty get a much larger percentual error if a positive and a negative bin are close to each other. It is therefore important that we put a constraint on how close the positive and negative bins multiples are allowed to be.

\subsection{Layer-specific findings}
We confirm the findings of \cite{cnn_noise} which claim that CNN's are more susceptible to noise on analog format. This was found by a larger loss of accuracy when drift was applied to the CNN layers compared to dense layers. 

We also find that there is inter-layer dependency between the layers given the type and amount of noise applied. aihwkit's \textbf{drift\_analog\_weights}-function drift weights equally if the same RPU-config is given. This means that often we will find that the layers drift in a similar stochastic fashion. This means that adjusting one single layer that has reached over a drift threshold $dt$ will often degrade the performance, as the inter-layer weight representation is dramatically changed instead of stochastically translated using the noise. This means that an approach where the entire network is re-adjusted can sometimes be better given the network and the conditions.

\subsection{Future research}

In order to access the methodology in practice, there needs to be an implementation of both techniques to hardware, and the result should be compared after periods of time.

A more robust approach would be to investigate the feasibility of an algorithm that combines the two methodologies, meaning that we do hardware-aware training whilst keeping the weights constrained close to multiples.

Another interesting area to explore is self-repair using bit-sliced network weights. This means that a network is represented with weights that are sliced into binary representations of 0s or 1s. This would make the weight adjustment scheme much more simple and flexible to various weights at the cost of more required hardware connections per weight.

Lastly, it would be interesting to see how the methodology performs on other types of analog memory architectures, such as RRAM which do not suffer from the same kinds of noise as PCM-based architectures.

\section{Conclusion}

We show that by using a constrained bin weight scheme, we can regain lost performance over time using a weight-multiple adjustment over a positive and negative part of the weight. We do however note that by not performing analog-aware training for PCM modules, the network becomes less stable. Despite regaining the accuracy back, the drift will affect the result between the resets more negatively than by using purely analog-awarely trained neural networks.

\clearpage  
\appendix
\section*{Appendix}
\section{Analog Noise Configuration}\label{code}

The following Python code snippet provides the configuration for the analog noise in the Phase-Change Memory (PCM) model. It sets up various parameters including weight noise, clip type, and drift compensation.

\begin{verbatim}
    rpu_config = InferenceRPUConfig()
    rpu_config.forward.out_res = -1.0  # Turn off (output) ADC discretization.
    rpu_config.forward.w_noise_type = WeightNoiseType.ADDITIVE_CONSTANT
    rpu_config.forward.w_noise = 0.02  # Short-term w-noise.

    rpu_config.clip.type = WeightClipType.FIXED_VALUE
    rpu_config.clip.fixed_value = 1.0
    rpu_config.modifier.pdrop = 0.03  # Drop connect.
    rpu_config.modifier.type = WeightModifierType.ADD_NORMAL  # Fwd/bwd weight noise.
    rpu_config.modifier.std_dev = 0.1
    rpu_config.modifier.rel_to_actual_wmax = True

    # Inference noise model.
    rpu_config.noise_model = PCMLikeNoiseModel(g_max=25.0)

    # drift compensation
    rpu_config.drift_compensation = GlobalDriftCompensation()
\end{verbatim}

\bibliographystyle{acmsiggraph}
\bibliography{project}
\end{document}